\documentclass[journal]{IEEEtran}

\usepackage{etoolbox}
\makeatletter
\@ifundefined{color@begingroup}%
{\let\color@begingroup\relax
\let\color@endgroup\relax}{}%
\def\fix@ieeecolor@hbox#1{%
\hbox{\color@begingroup#1\color@endgroup}}
\patchcmd\@makecaption{\hbox}{\fix@ieeecolor@hbox}{}{\FAILED}
\patchcmd\@makecaption{\hbox}{\fix@ieeecolor@hbox}{}{\FAILED}

\usepackage{cite}
\usepackage{amsmath,amssymb,amsfonts}
\usepackage{algorithm}
\usepackage{graphicx}
\usepackage{textcomp}

\usepackage{array}
\usepackage{braket}
\usepackage{algpseudocode}
\usepackage{booktabs}
\usepackage{placeins}
\usepackage{stfloats}
\usepackage{url}
\usepackage{verbatim}

\def\BibTeX{{\rm B\kern-.05em{\sc i\kern-.025em b}\kern-.08em
    T\kern-.1667em\lower.7ex\hbox{E}\kern-.125emX}}
\begin{document}
\title{QuantEIT: Ultra-Lightweight Quantum-Assisted Inference for Chest Electrical Impedance Tomography}
\author{Hao Fang, \IEEEmembership{Graduate Student member, IEEE}, Sihao Teng, \IEEEmembership{Graduate Student member, IEEE}, Hao Yu, \IEEEmembership{Member, IEEE}, Siyi Yuan, Huaiwu He, Zhe Liu, \IEEEmembership{Member, IEEE}, Yunjie Yang, \IEEEmembership{Senior Member, IEEE}
\thanks{Hao Fang, Sihao Teng, Hao Yu, Zhe Liu and Yunjie Yang are with the SMART Group, Institute for Imaging, Data and Communications (IDCOM), School of Engineering, The University of Edinburgh, Edinburgh, UK. (Correspondence authors: Yunjie Yang and Zhe Liu; Email: y.yang@ed.ac.uk and zz.liu@ed.ac.uk). }
\thanks{Siyi Yuan and Huaiwu He are with the State Key Laboratory of Complex Severe and Rare Diseases, Department of Critical Care Medicine, Peking Union Medical College, Peking Union Medical College Hospital, Chinese Academy of Medical Sciences, Beijing, China. }
\thanks{This work was supported in part by Noncommunicable Chronic Diseases-National Science and Technology Major Project (2024ZD0522700), and in part by Prunus Medical-Edinburgh joint research grant.}}

\maketitle

\begin{abstract}
Electrical Impedance Tomography (EIT) is a non-invasive, low-cost bedside imaging modality with high temporal resolution, making it suitable for bedside monitoring. However, its inherently ill-posed inverse problem poses significant challenges for accurate image reconstruction. Deep learning (DL)-based approaches have shown promise but often rely on complex network architectures with a large number of parameters, limiting efficiency and scalability. Here, we propose an Ultra-Lightweight Quantum-Assisted Inference (QuantEIT) framework for EIT image reconstruction. QuantEIT leverages a Quantum-Assisted Network (QA-Net), combining parallel 2-qubit quantum circuits to generate expressive latent representations that serve as implicit nonlinear priors, followed by a single linear layer for conductivity reconstruction. This design drastically reduces model complexity and parameter number. Uniquely, QuantEIT operates in an unsupervised, training-data-free manner and represents the first integration of quantum circuits into EIT image reconstruction. Extensive experiments on simulated and real-world 2D and 3D EIT lung imaging data demonstrate that QuantEIT outperforms conventional methods, achieving comparable or superior reconstruction accuracy using only \textit{0.2\%} of the parameters, with enhanced robustness to noise.
\end{abstract}


\section{Introduction}
\label{sec:introduction}
\IEEEPARstart{E}{lectrical} Impedance Tomography (EIT) is a non-invasive imaging technique that reconstructs spatially resolved conductivity distributions by injecting small electrical currents and measuring the resulting boundary voltages \cite{b2,b1}. Due to its non-ionizing nature, low cost, and high temporal resolution, EIT has been widely applied in medical imaging (e.g., pulmonary monitoring \cite{b3,b4,b5} and brain function imaging \cite{b6,b7,b8}) and industrial process monitoring \cite{b9,b10}. However, the inherently ill-posed nature of EIT fundamentally limits its spatial resolution and image quality, posing substantial challenges for accurate and robust image reconstruction. 

To address such issues, traditional reconstruction methods have relied on handcrafted regularization techniques such as Tikhonov \cite{b11}, Newton's One-Step Error Reconstructor (Noser) \cite{b22}, $\ell_1$ norms \cite{b12}, and Total Variation (TV) regularization \cite{b13}. While these approaches help stabilize the inversion process, they often require careful parameter tuning and tend to perform poorly in severely ill-posed or noisy scenarios. In the past decade, supervised deep learning (DL)-based methods \cite{b14,b16,b38,b46} have been extensively explored for inverse problems in imaging, including EIT. Compared to handcrafted priors, these methods, including convolutional neural networks (CNNs) \cite{b23,b24,b25}, variational autoencoders (VAEs) \cite{b26}, and generative adversarial networks (GANs) \cite{b27}, have demonstrated notable improvements by learning complex features and data-driven priors from large labeled datasets. However, such approaches heavily rely on the availability of large-scale, high-quality labeled datasets, which can be difficult and costly to obtain in practical EIT settings. Moreover, these models are typically task-specific and often exhibit limited generalization to unseen structures, patterns, or imaging conditions \cite{b30,b31}.

To overcome the limitations of supervised learning, unsupervised DL-based methods, exemplified by Deep Image Prior (DIP) \cite{b21}, have attracted attention for their training-data-free nature and strong generalization capability. These approaches use untrained neural networks (UNNs) as implicit regularizers, representing the reconstructed conductivity distribution directly as the output of the network. Leveraging the implicit regularization of the network itself, reconstruction is performed by iteratively updating its parameters. By optimizing the network parameters using only the measurement data, these methods avoid the need for pretraining on large datasets. D. Liu et al. \cite{b28} introduced DIP to EIT by employing a U-Net as an implicit regularizer, reformulating the reconstruction as a neural network parameter optimization problem and achieving stable and high-quality 2D EIT reconstructions. Later, Z. Liu et al. \cite{b29} proposed Regularized Shallow Image Prior (R-SIP), incorporating handcrafted regularization into a shallow three-layer multilayer perceptron (MLP), achieving robust reconstruction in both 2D and 3D settings. Fang et al. \cite{b30}, on the other hand, presented Multi-Branch Attention Image Prior (MAIP), an unsupervised framework enhancing generalization and enabling robust 2D multifrequency EIT reconstruction.

However, these unsupervised methods still rely on conventional neural networks, which approximate complex nonlinear mappings through stacked layers of linear transformations and nonlinear activations, typically requiring carefully designed architectures with numerous parameters. As a result, these models can be computationally intensive, prone to overfitting, and difficult to generalize across imaging scenarios. Moreover, the need for extensive tuning and high memory usage poses challenges for real-time or lightweight deployment.

To address these challenges, recent studies have explored integrating quantum circuits \cite{b32} into neural architectures to form hybrid quantum–classical models. Quantum circuits inherently offer high expressive power with surprisingly few parameters, enabled by uniquely quantum phenomena like superposition and entanglement \cite{b33,b34}. These capabilities allow quantum-assisted models to represent complex nonlinear mappings more compactly and efficiently than conventional deep neural networks. For example, quantum convolutional neural networks (QCNNs) have achieved high image-classification accuracy despite using only a limited number of trainable parameters, highlighting their capacity to capture rich feature representations with a compact model size \cite{b35}.

These advantages open up new possibilities for building lightweight, generalizable quantum-assisted inference frameworks to tackle ill-posed inverse problems like EIT, particularly under data-scarce or resource-constrained scenarios. In particular, quantum models have demonstrated promising generalization from remarkably small training sets. One theoretical analysis guarantees, for instance, that a QCNN with only $O(\log n)$ parameters can learn to classify quantum states using just $O((\log n)^2)$ training samples, suggesting robust performance even in severely data-limited scenarios. Here, $n$ denotes the number of qubits representing the input state, and the logarithmic scaling indicates that both the parameter count and the sample complexity grow slowly with system size \cite{b36}.

Here, we propose an Ultra-Lightweight Quantum-Assisted Iterative Inference (QuantEIT) framework for EIT image reconstruction. We introduce parallel quantum circuits composed of parameterized single-qubit rotation gates and two-qubit entanglement gates to generate compact latent representations for EIT image reconstruction. By leveraging the expressive power of quantum circuits, our method requires only a single linear mapping layer to achieve high-quality 2D and 3D reconstructions. This design effectively avoids designing and tuning complex network architectures while significantly reducing the number of model parameters. In terms of reconstruction performance, the framework also exhibits strong robustness to measurement noise and generalizes effectively across different anatomical structures and imaging conditions.

Our main contributions are summarized as follows:
\begin{itemize}
    \item We propose QuantEIT, a novel ultra-lightweight quantum-assisted iterative inference framework that integrates parallel quantum circuits into EIT reconstruction, drastically reducing model complexity for efficient inference.
    
    \item QuantEIT enables high-quality 2D and 3D reconstructions in an unsupervised manner. It employs a compact Quantum-Assisted Network (QA-Net) that implicitly generates latent representations from fixed trainable quantum parameters. This module, composed of two parallel 2-qubit circuits and a single linear layer, serves as a lightweight implicit nonlinear prior for the inverse mapping. 

    \item Comprehensive experiments on simulated and real-world data show that QuantEIT outperforms state-of-the-art classical models, achieving superior reconstruction quality with only \textit{0.2\%} of the parameters and enhanced robustness to noise.
\end{itemize}

\section{Methodology}
\begin{figure*}
\centerline{\includegraphics[scale=0.35]{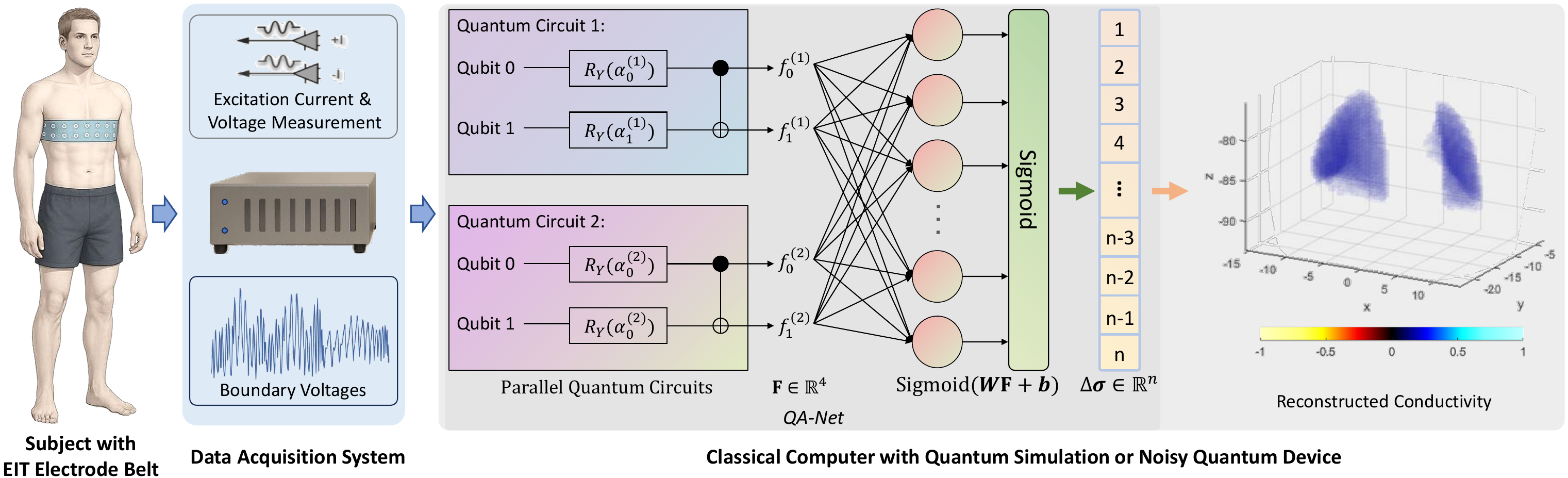}}
	\caption{Overview of the hybrid quantum-classical framework for EIT imaging.}
	\label{flowchart}
\end{figure*}

Fig.~\ref{flowchart} shows the hybrid quantum–classical framework for chest impedance imaging. Currents are first injected into the body through an electrode belt, and boundary voltages are measured by the data acquisition system. The recorded voltages are then processed on a classical computer, which by default interfaces with quantum circuits simulated locally using the PennyLane \cite{b37} framework. Alternatively, these circuits can also be executed on a quantum backend. The quantum circuits generate expressive latent representations that are mapped to conductivity distributions. 

\subsection{EIT image reconstruction}
EIT image reconstruction is a typical inverse problem that aims to reconstruct 2D or 3D conductivity distributions from voltage measurements, typically subject to noise. The EIT forward model, which characterizes the nonlinear relationship between internal conductivity changes and boundary voltage responses, can be linearized around a known reference conductivity, yielding:
\begin{equation}
\Delta\mathbf{v} = \mathbf{J\Delta\pmb{\sigma}},
\label{e0}
\end{equation}
where $\Delta\mathbf{v}=(\mathbf{v_{o}}-\mathbf{v_{r}})\oslash\mathbf{v_{r}}\in\mathbb{R}^{m}$ and $\Delta\pmb{\sigma}=-(\mathbf{\pmb{\sigma}_{o}}-\mathbf{\pmb{\sigma}_{r}})\oslash\mathbf{\pmb{\sigma}_{r}}\in\mathbb{R}^{n}$ denote the normalized voltage measurements and conductivity distributions, respectively. $\mathbf{J}\in\mathbb{R}^{m \times n}$ represents the Jacobian (or sensitivity) tensor. Here, $m$ is the number of independent voltage measurements, and $n$ is the number of elements (pixels or voxels). 

Due to the ill-posed nature of EIT, directly solving \eqref{e0} is highly underdetermined and noise-sensitive. To mitigate this issue, we reformulate the reconstruction problem as a regularized optimization task. Specifically, the normalized conductivity change $\Delta\pmb{\sigma}^{*}$ is estimated by solving:
\begin{align}
\Delta\pmb{\sigma}^{*} = \arg\min_{\Delta\pmb{\sigma}} \quad 
& \bigl|\Delta\mathbf{v} - \mathbf{J}\,\Delta\pmb{\sigma}\bigr| \notag + \pmb{\lambda}^{\top}\,\pmb{\mathcal{R}}\bigl(\Delta\pmb{\sigma}\bigr),
\label{eq:argmin_multi_reg}
\end{align}
where $|\cdot|$ denotes the selected data fidelity norm (e.g., $\ell_1$ or $\ell_2$). 
$\pmb{\mathcal{R}}(\cdot)$ is a vector of handcrafted regularization terms imposing different prior constraints on the solution, and $\pmb{\lambda}^{\top}$ is the corresponding vector of regularization weights that control the influence of each prior term.

\subsection{Quantum Circuit}
A quantum bit (qubit) resides in a 2D complex Hilbert space \cite{b41}. The standard computational basis vectors are denoted as:
\begin{equation}
\ket{0} = \begin{pmatrix}1\\0\end{pmatrix}, \quad \ket{1} = \begin{pmatrix}0\\1\end{pmatrix},
\end{equation}

For a system of \( n_q=2 \) qubits, the total Hilbert space is \( \mathcal{H} = \mathbb{C}^{2^{n_q}}=\mathbb{C}^{4} \), with the standard basis spanning states like:
\begin{equation}
\{\ket{00}, \ket{01}, \ket{10}, \ket{11}\},
\end{equation}

A valid quantum state \( \ket{\psi} \in \mathcal{H} \) must satisfy the normalization condition:
\begin{equation}
\braket{\psi | \psi} = 1,
\end{equation}
which ensures that the total probability of all possible measurement outcomes is one.

The first stage of circuit evolution applies parameterized single-qubit rotation gates around the Y-axis, described by the matrix:
\begin{equation}
R_Y(\alpha) = \exp\left(-i \frac{\alpha}{2} Y\right) = \begin{pmatrix}
\cos\frac{\alpha}{2} & -\sin\frac{\alpha}{2} \\
\sin\frac{\alpha}{2} & \cos\frac{\alpha}{2}
\end{pmatrix},
\label{e1}
\end{equation}

The Pauli-Y operator is defined as:
\begin{equation}
Y = \begin{pmatrix}
0 & -i \\
i & 0
\end{pmatrix}.
\end{equation}

Applying \( R_Y(\alpha) \) to the ground state \( \ket{0} \) yields:
\begin{equation}
R_Y(\alpha) \ket{0} = \cos\frac{\alpha}{2} \ket{0} + \sin\frac{\alpha}{2} \ket{1}.
\end{equation}

This unitary and real-valued transformation maps each qubit into a continuously differentiable superposition, enabling analytic gradient computation via the parameter-shift rule.

To introduce entanglement, a chain of Controlled-NOT (CNOT) gates is applied. A CNOT gate acts on two qubits (control \( c \), target \( t \)) and transforms the joint state as:
\begin{equation}
\ket{c, t} \mapsto \ket{c} \otimes \ket{t \oplus c},
\end{equation}

A CNOT gate flips the target qubit if and only if the control qubit is \( \ket{1} \). The standard 2-qubit CNOT matrix is:
\begin{equation}
\text{CNOT} = \begin{pmatrix}
1 & 0 & 0 & 0 \\
0 & 1 & 0 & 0 \\
0 & 0 & 0 & 1 \\
0 & 0 & 1 & 0
\end{pmatrix}.
\label{e2}
\end{equation}

This layer induces entanglement between adjacent qubits, significantly enhancing the expressive power of the circuit.

After rotations and entanglement, the final quantum state \( \ket{\psi(\pmb{\phi})} \) is measured. For each qubit \( k \), the expectation value of the Pauli-Z operator is computed as:
\begin{equation}
\begin{aligned}
f_k(\pmb{\phi}) &= \braket{\psi(\pmb{\phi}) | Z_k | \psi(\pmb{\phi})}, \\
Z_k &= I^{\otimes k} \otimes Z \otimes I^{\otimes (n_q - k - 1)}, Z = \begin{pmatrix} 1 & 0 \\ 0 & -1 \end{pmatrix}.
\end{aligned}
\label{e3}
\end{equation}

This corresponds to measuring the \( k \)-th qubit in the computational basis, assigning \( +1 \) to \( \ket{0} \) and \( -1 \) to \( \ket{1} \).

For an independent quantum circuit with \( n_q \) qubits indexed by $i$ , this results in a real-valued feature vector:
\begin{equation}
\mathbf{f}^{(i)} = [f_0^{(i)}, f_1^{(i)}, \ldots, f_{n_q-1}^{(i)}]^\top \in \mathbb{R}^{n_q}.
\end{equation}

This vector encodes the response of the quantum circuit to the parameterized gates.

To scale this mechanism, multiple independent quantum circuits (indexed by \( i = 1, \ldots, n_c \)) are instantiated in parallel. Each circuit has its own parameters \( \pmb{\phi}^{(i)} \) and produces a feature vector \( \mathbf{f}^{(i)} \in \mathbb{R}^{n_q} \). These are concatenated to yield a unified feature:
\begin{equation}
\mathbf{F} = [\mathbf{f}^{(1)}, \mathbf{f}^{(2)}, \ldots, \mathbf{f}^{(n_c)}]^\top \in \mathbb{R}^{n_q \cdot n_c}.
\label{e4}
\end{equation}

This compact and expressive vector serves as the input to a subsequent classical neural network layer. During training, all parameters $\pmb{\phi}^{(i)}$ are differentiable, and their gradients $\frac{\partial f_k}{\partial \pmb{\phi}_j}$ can be computed using automatic differentiation frameworks such as PennyLane \cite{b37}, enabling seamless integration into classical backpropagation pipelines. Gradient descent optimization, such as Adam \cite{b40}, is used to update parameters across all circuits in a unified training loop.



\subsection{QA-Net for Implicit Latent Generation}
The \emph{QA-Net} consists of $n_c$ parallel quantum circuits, each operating on $n_q$ qubits, followed by a classical linear mapping layer. It defines a function $\Psi: \pmb{\phi} \to \mathbb{R}^{n}$, where $\pmb{\phi} \in \mathbb{R}^{n_c \cdot n_q}$ denotes the set of trainable quantum parameters. Increasing \( n_q \) enhances the expressive power of each circuit by expanding its entangled Hilbert space, while increasing \( n_c \) augments the overall latent dimension, enabling the model to represent a broader class of solutions with greater structural diversity. In this study, we set $n_q = 2, n_c = 2$ to form a minimal quantum-assisted module, balancing expressivity with hardware efficiency, as illustrated in Fig.~\ref{flowchart}. These circuits do not require explicit input but instead operate through parameterized quantum gates to produce a compact latent representation \( \mathbf{F} \in \mathbb{R}^{4} \). This latent vector is then passed to a classical linear layer, which maps it to the one-dimensional conductivity distribution \( \Delta\pmb{\sigma} \in \mathbb{R}^{n} \). The mapping is defined as:
\begin{equation}
\Delta\pmb{\sigma} =  \mathrm{Sigmoid}(\pmb{W} \mathbf{F} + \pmb{b}),
\end{equation}
where \( \pmb{W} \in \mathbb{R}^{n \times 4} \) and \( \pmb{b} \in \mathbb{R}^n \) are learnable parameters, and \( \mathrm{Sigmoid}(\cdot) \) denotes the element-wise logistic activation function, ensuring that the output values of \( \Delta\pmb{\sigma} \) lie within the normalized range \([0, 1]\).

In this hybrid architecture, the quantum circuits act as expressive nonlinear mappings that implicitly regularize the reconstruction, enabling lightweight, training-free inference for solving ill-posed inverse problems such as EIT.

\subsection{QuantEIT for EIT image reconstruction}
Within our QuantEIT framework, the unknown conductivity distribution is expressed as the output of the proposed hybrid quantum-classical \emph{QA-Net}, denoted as $\Psi (\pmb{\theta})$, where $\pmb{\theta}= (\pmb{\phi},\pmb{W},\pmb{b})$, i.e.,
\begin{equation}
\Delta\pmb{\sigma}^{(k)}=\Psi\left(\pmb{\phi}^{(k)},\pmb{W}^{(k)}, \pmb{b}^{(k)}\right)=\Psi\left(\pmb{\theta}^{(k)}\right),
\label{eq:QAEIT}
\end{equation}
where \( k \) denotes the optimization step in the iterative reconstruction process. To fit the predicted voltages to the measured ones, the parameters \( \pmb{\theta} \) are optimized by minimizing the following regularized loss:
\begin{equation}
\mathcal{L}(\pmb{\theta}) = 
\left\| \Delta \mathbf{v} - \mathbf{J}\,\Psi(\pmb{\theta}) \right\|^2 
+ 
\pmb{\lambda}^{\top}\,\pmb{\mathcal{R}}\bigl(\Psi(\pmb{\theta})\bigr),
\label{eq:loss_multi_reg}
\end{equation}
In our experiments, to further mitigate the ill-posedness inherent in EIT, we selected three commonly used and complementary handcrafted regularization terms—Laplacian smoothing~\cite{b43}, Total Variation (TV)~\cite{b42}, and $\ell_1$ norm~\cite{b12}. These regularizers provide synergistic constraints on different structural features, enhancing the robustness of our unsupervised reconstruction method under uncertainty. Accordingly, the regularization weight vector and the regularization function vector are defined as:
\[
\pmb{\lambda} = 
\begin{bmatrix}
\lambda_{\mathrm{Laplacian}} \\[3pt]
\lambda_{\mathrm{TV}} \\[3pt]
\lambda_{\ell_1}
\end{bmatrix},
\quad
\pmb{\mathcal{R}}\bigl(\Psi(\pmb{\theta})\bigr) = 
\begin{bmatrix}
\mathcal{R}_{\mathrm{Laplacian}}\bigl(\Psi(\pmb{\theta})\bigr) \\[3pt]
\mathcal{R}_{\mathrm{TV}}\bigl(\Psi(\pmb{\theta})\bigr) \\[3pt]
\mathcal{R}_{\ell_1}\bigl(\Psi(\pmb{\theta})\bigr)
\end{bmatrix}.
\]

Accordingly, the model parameters at each iteration are updated as:
\begin{equation}
\pmb{\theta}^{(k+1)} = \arg\min_{\pmb{\theta}} \mathcal{L}(\pmb{\theta}),
\label{eq:adam}
\end{equation}
where the optimization is performed using the Adam optimizer, which adaptively adjusts learning rates based on the first and second moments of the gradients.

After $N$ iterations, the final conductivity distribution is obtained by evaluating the model at the last parameter state:
\begin{equation}
\Delta\pmb{\sigma}^{*} = \Psi(\pmb{\theta}^{(N)}).
\label{eq:final_sigma}
\end{equation}

\begin{algorithm}[t]
\caption{QuantEIT for EIT Reconstruction}
\begin{algorithmic}[1]
\Require Normalized voltage $\Delta\mathbf{v}$, sensitivity matrix $\mathbf{J}$, regularization weights $\pmb{\lambda}$, number of iterations $N$
\State Initialize model parameters $\pmb{\theta}^{(0)} = (\pmb{\phi}^{(0)}, \pmb{W}^{(0)}, \pmb{b}^{(0)})$
\For{$k = 0$ to $N-1$}
    \State Compute conductivity update: $\Delta\pmb{\sigma}^{(k)} = \Psi(\pmb{\theta}^{(k)})$
    \State Compute loss: 
    \[
    \mathcal{L}\bigl(\pmb{\theta}^{(k)}\bigr) = \left\| \Delta\mathbf{v} - \mathbf{J}\,\Delta\pmb{\sigma}^{(k)} \right\|^2 
    + 
    \pmb{\lambda}^{\top}\,\pmb{\mathcal{R}}\bigl(\Delta\pmb{\sigma}^{(k)}\bigr)
    \]
    \State Update parameters via Adam:
    \[
    \pmb{\theta}^{(k+1)} = \text{Adam}\bigl(\pmb{\theta}^{(k)}, \nabla_{\pmb{\theta}}\,\mathcal{L}\bigl(\pmb{\theta}^{(k)}\bigr)\bigr)
    \]
\EndFor
\Ensure Reconstructed conductivity distribution $\Delta\pmb{\sigma}^{*} = \Psi\bigl(\pmb{\theta}^{(N)}\bigr)$
\end{algorithmic}
\end{algorithm}

\section{Experimental Setup}
\subsection{Simulation Data Generation}
\begin{figure}
\centerline{\includegraphics[scale=0.25]{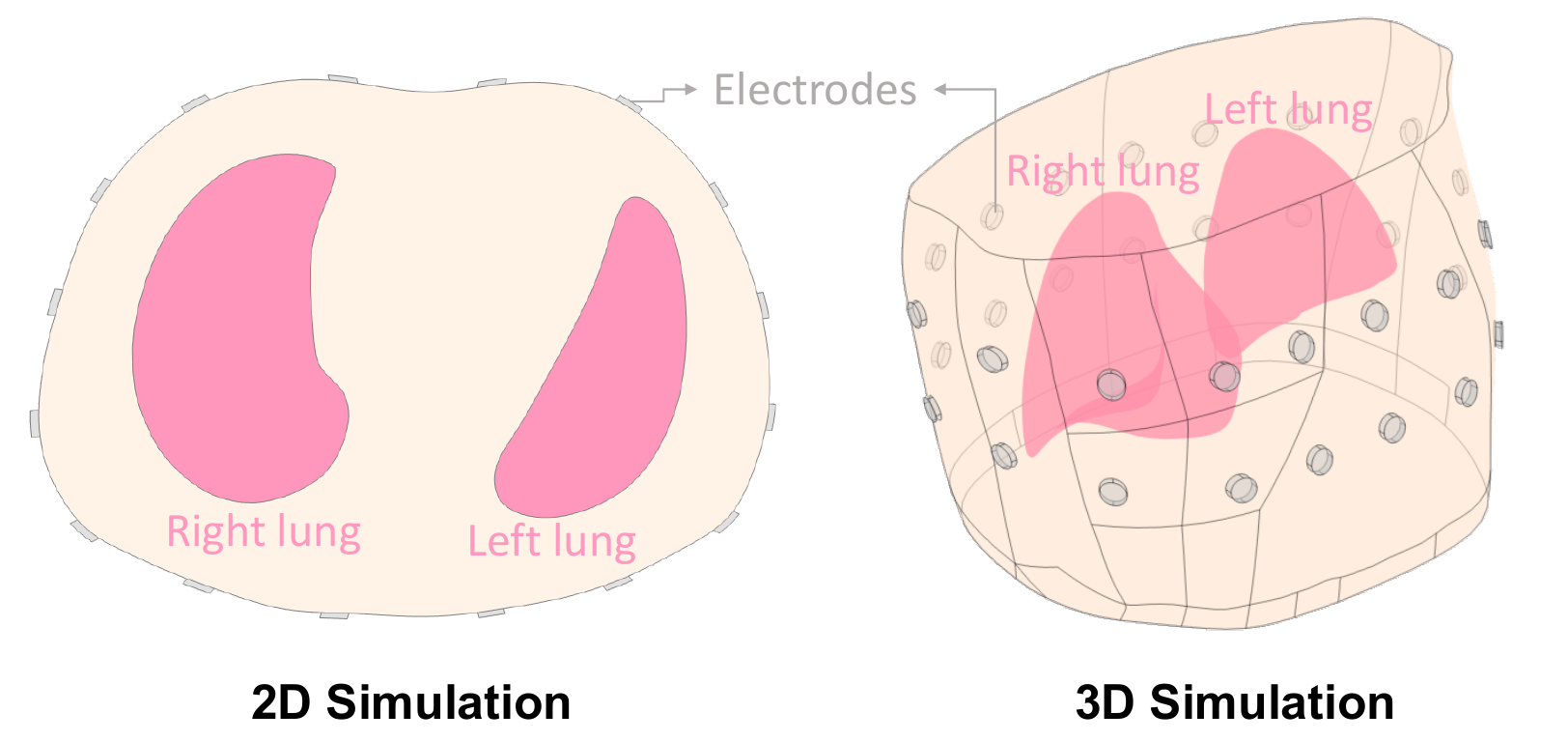}}
	\caption{Illustration of the 2D and 3D simulation setups. The left shows the 2D thoracic cross-section with embedded lung regions and boundary electrodes. The right depicts the 3D thoracic model with two circumferential electrode layers and the internal lung structures.}
	\label{simulation_exp}
\end{figure}

\begin{figure}
\centerline{\includegraphics[scale=0.3]{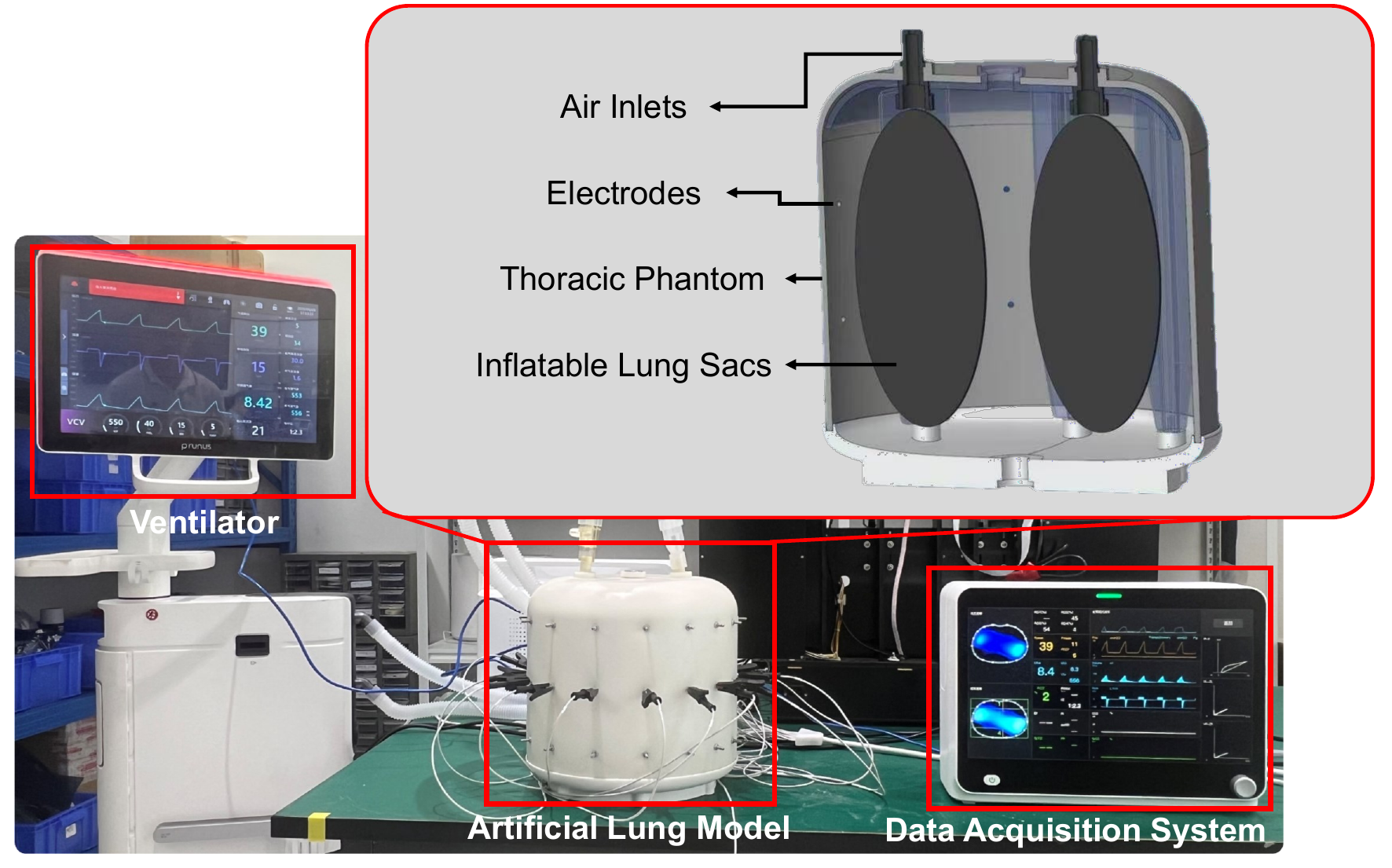}}
	\caption{Illustration of the lung phantom experiment setup. }
	\label{phantom_exp}
\end{figure}

\begin{figure}
\centerline{\includegraphics[scale=0.36]{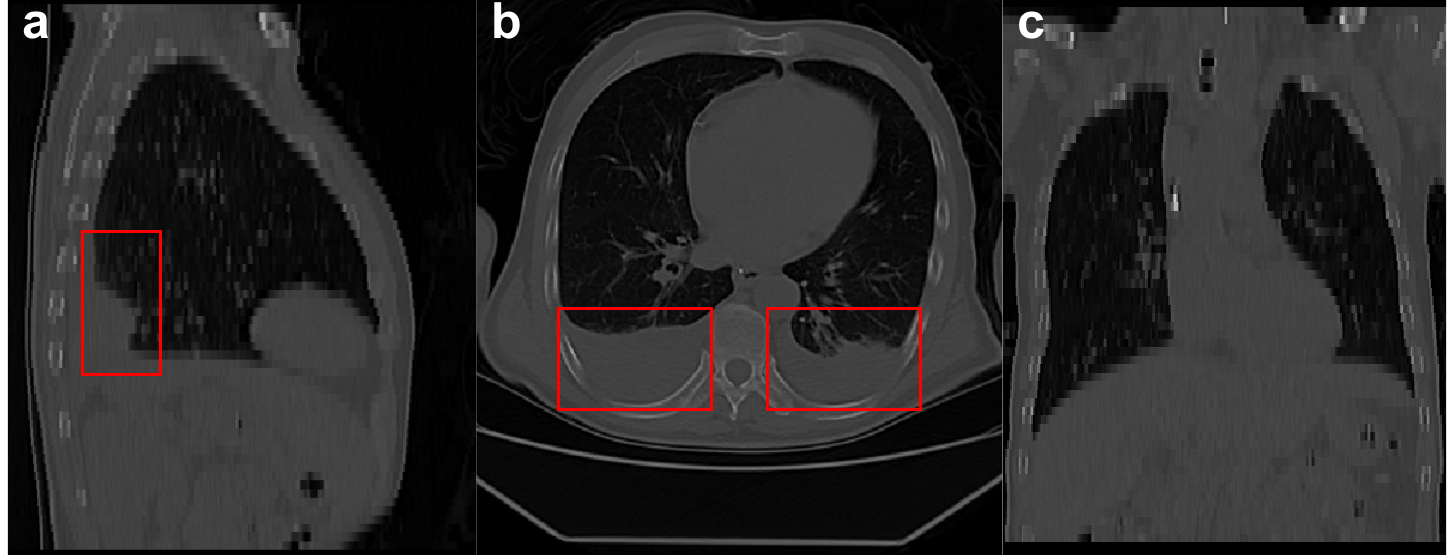}}
	\caption{CT scans of Participant 2. (a) Sagittal view, (b) Axial view, and (c) Coronal view. The red boxes indicate dorsal lung volume loss consistent with posterior atelectasis.}
	\label{CT_clinical}
\end{figure}

\begin{figure}
\centerline{\includegraphics[scale=0.38]{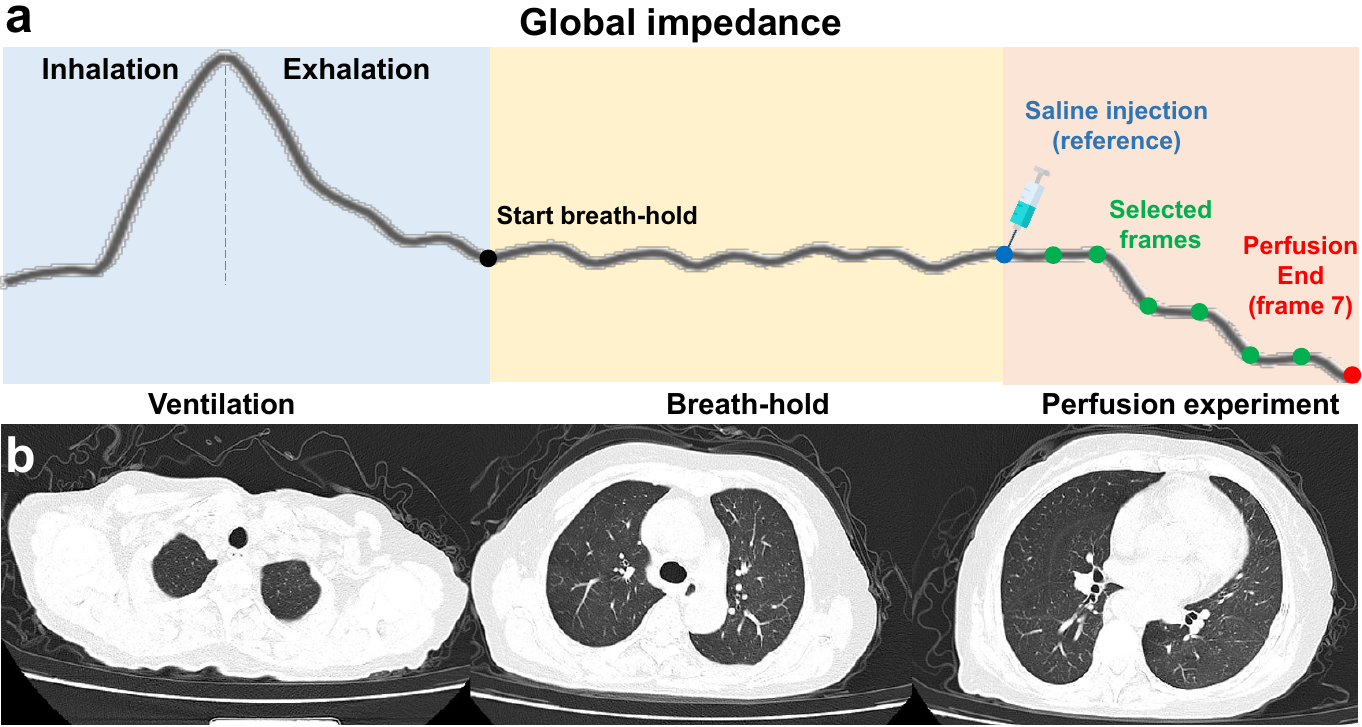}}
	\caption{Global impedance dynamics and thoracic CT images from Participant 3. (a) The impedance curve shows ventilation, breath hold, and perfusion phases. Key events—including the start of breath-hold, saline injection (reference), selected perfusion frames, and perfusion end—are marked with dots. (b) Representative axial CT slices.}

	\label{CT_clinical2}
\end{figure}
We employed COMSOL Multiphysics (v6.1) to simulate 2D and 3D EIT measurements using anatomically realistic thoracic models derived from CT scans, as illustrated in Fig.~\ref{simulation_exp}. In the 2D setup, 16 electrodes were uniformly arranged along the boundary of the thoracic cross-section, with the lung regions positioned centrally. In the 3D setup, 32 electrodes were evenly distributed in two circumferential layers enclosing the 3D lung geometry. 

The background conductivity was set to \(0.24~\mathrm{S/m}\) in both the 2D and 3D simulation cases. In the 2D model, the left lung conductivity was assigned \(0.17~\mathrm{S/m}\) and the right lung \(0.14~\mathrm{S/m}\), simulating a scenario of left lung ventilation impairment. This configuration was designed to evaluate whether different reconstruction methods can accurately distinguish multiple objects with varying conductivity levels within the same imaging domain. In the 3D model, both lungs were set to \(0.20~\mathrm{S/m}\), representing the early phase of inhalation, to assess the capability of each method to reconstruct subtle conductivity changes. A total of 104 boundary voltage measurements were generated for the 2D case and 328 measurements for the 3D case, following the measurement protocol in \cite{b29}.

\subsection{Lung Phantom Experiment Setup}

Fig.~\ref{phantom_exp} shows the lung phantom experiment setup. A custom-designed thoracic phantom was built to emulate lung ventilation dynamics, consisting of two inflatable lung sacs enclosed within a cylindrical housing and surrounded by evenly distributed electrodes. Air inlets on the top enable independent control of inflation via an external ventilator. Fig. 3(b) shows the real-world experimental setup, including a commercial mechanical ventilator, an artificial lung model, and an EIT system (Shenzhen Yuanlu Inc., EIT V100).

We simulated unilateral pulmonary atelectasis by ventilating only the left lung sac while keeping the right sac uninflated. Voltage measurements were acquired using the same measurement paradigm as in the 3D simulation. The frame prior to ventilation was selected as the reference, while the frame captured at the onset of left lung inflation—when the left sac just begins to expand—was used as the observation. This setup offers a controlled yet realistic scenario to evaluate the reconstruction performance of different algorithms. Specifically, the reference-to-observation transition involves only minor conductivity changes, simulating a weak signal condition that challenges the algorithm's ability to detect subtle physiological variations under realistic noise and uncertainties.

\subsection{Clinical Data Acquisition}

Clinical data were acquired through real-world experiments. The study protocol was approved by the Institutional Review Board (IRB) of Peking Union Medical College Hospital (approval number: I-24PJ1732). Prior to any data collection, written informed consent was obtained from participants using an IRB-approved consent form (version 2.0, dated July 22, 2024). All procedures complied with the Declaration of Helsinki and national ethical regulations. Participant data were anonymized to ensure confidentiality.

Three participants were enrolled. Participant 1 underwent 2D measurements using a 16-electrode belt. This participant had left lung ventilation impairment, consistent with the 2D simulation condition. Participants 2 and 3 underwent 3D EIT measurements using a dual-layer 32-electrode belt. Participants 1 and 2 underwent ventilation-only experiments, while Participant 3 underwent both ventilation and perfusion procedures. Throughout all procedures, participants remained in a semi-recumbent position and followed a controlled breathing protocol. All EIT voltage data were acquired using commercial systems (Shenzhen Yuanlu Inc.; V100 for 2D, V200 for 3D) following the same protocol as used in the simulation experiments. CT scans were also acquired for Participant 2 and Participant 3 to provide anatomical references for subsequent analysis (see Fig.~\ref{CT_clinical} and Fig.~\ref{CT_clinical2}b).

For Participants 1 and 2, the voltage measurements acquired at the start of inhalation were used as the reference. For Participant 1, the observation voltage was taken at the end of inhalation, capturing the full ventilation-induced conductivity change. For Participant 2, the observation voltage corresponded to the early phase of the breathing cycle, when the lungs had just begun to expand. 

For Participant 3, the reference voltage was selected at the onset of pulmonary perfusion, as shown in Fig.~\ref{CT_clinical2}a. Prior to the experiment, standard pre-checks were performed to confirm the patency of the central venous catheter, and a 5\% hypertonic saline solution was prepared by mixing 10\,ml of sterile water with 10\,ml of 10\% NaCl. During the procedure, an end-expiratory breath-hold maneuver was initiated under deep sedation to suppress spontaneous respiration (breath-hold duration $<8\,s$). Once a stable plateau appeared in the global impedance waveform, 10\,ml of the prepared saline was rapidly injected via the central line within 2\,seconds. EIT data acquisition was simultaneously triggered to capture the impedance dynamics during the first-pass circulation of the contrast agent. A sequence of seven consecutive voltage frames was then selected to span the entire perfusion process for time-sequence reconstruction.

\subsection{Comparison Methods}
We compare QuantEIT with two baselines. The first is Newton’s One-Step Error Reconstructor (Noser) \cite{b22}, commonly used in clinical settings \cite{b44}. The second is Regularized Shallow Image Prior (R-SIP) \cite{b29}, the state-of-the-art unsupervised learning approach for 2D and 3D EIT. R-SIP combines handcrafted regularization with shallow neural networks to enhance reconstruction robustness and performance. We set the Noser parameter \(\mu\) to 20 based on trial and error. For R-SIP, we adopted the optimal parameter configurations reported in \cite{b29} to ensure fair comparisons.

\subsection{Implementation Details}
We set both \(n_q\) (the number of qubits per circuit) and \(n_c\) (the number of parallel quantum circuits) to 2 to construct an ultra-lightweight quantum-assisted network. Notably, choosing \(n_q=2\) corresponds to the minimal configuration required to introduce entanglement between qubits. This configuration provides a compact yet expressive latent space while maintaining compatibility with near-term quantum hardware.

For all simulation and real-world experiments, the number of optimization iterations was fixed to 1000. The learning rate was set to 0.05 for the 2D simulation and real-world experiments, while a higher value of 0.1 was employed for the 3D simulation and real-world experiments based on convergence curve analysis under different learning rates. For the 2D simulation and real-world experiments, the regularization weight vector \(\pmb{\lambda}^{\top}\) was set to \([0.03,\,0.002,\,0.01]\) and \([0.05,\,0.03,\,0.1]\), respectively, based on trial and error. For the 3D simulation and real-world experiments, \(\pmb{\lambda}^{\top}\) was fixed at \([0.001,\,0.001,\,0.001]\).

\section{Results and Discussion}
\subsection{Simulation Experiments}
\subsubsection{Convergence and Noise Resistance}



\begin{figure}
\centerline{\includegraphics[scale=0.4]{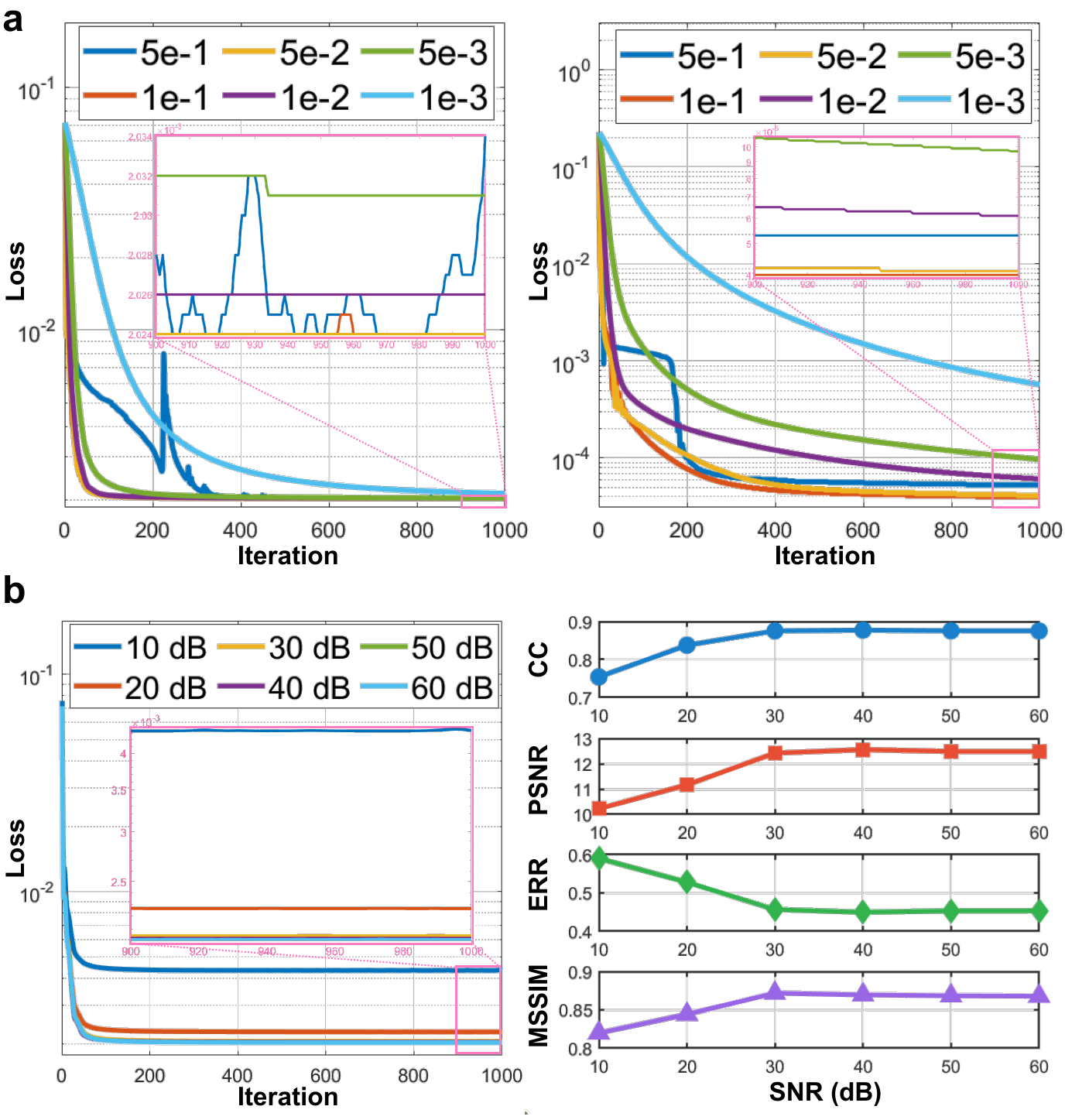}}
\caption{Convergence and robustness analysis of QuantEIT.  
(a) Convergence under different learning rates for 2D (left) and 3D (right) simulations.  
(b) Effect of noise level on 2D simulation: loss curves under different SNRs (left), and quantitative metrics (CC, PSNR, ERR, MSSIM) versus SNR (right).}

	\label{curves}
\end{figure}

We evaluated the convergence behavior of QuantEIT using both 2D and 3D simulation measurements (see Figure~\ref{curves}a). Specifically, we performed identical reconstruction tasks under different learning rates and plotted the loss curves over iterations to identify the optimal learning rates. We observed that excessively large learning rates led to oscillations in the loss, while overly small learning rates resulted in slow convergence. Based on the convergence curves, we determined that a learning rate of 0.05 for the 2D scenario and 0.1 for the 3D scenario yielded the fastest and smoothest convergence.

We further investigated the robustness of QuantEIT to measurement noise using 2D simulation data. We introduced additive Gaussian noise to the voltage inputs to simulate signal-to-noise ratios (SNRs) ranging from 10 dB to 60 dB. As shown in Figure~\ref{curves}b, the left panel plots the loss curves during optimization, which converge smoothly across all SNR levels. The right panel presents four quantitative metrics as functions of SNR: Correlation Coefficient (CC), Peak Signal-to-Noise Ratio (PSNR), Relative Error (ERR)\cite{b29}, and Mean Structural Similarity Index (MSSIM)~\cite{b45}. The results indicate that QuantEIT maintains high reconstruction quality for SNRs above 30 dB. Even under strong noise conditions (10–20 dB), the performance degradation remains within approximately 10\%, demonstrating the strong noise resilience of our method.

\subsubsection{Qualitative and Quantitative Reconstruction Results}

\begin{figure}
\centerline{\includegraphics[scale=0.37]{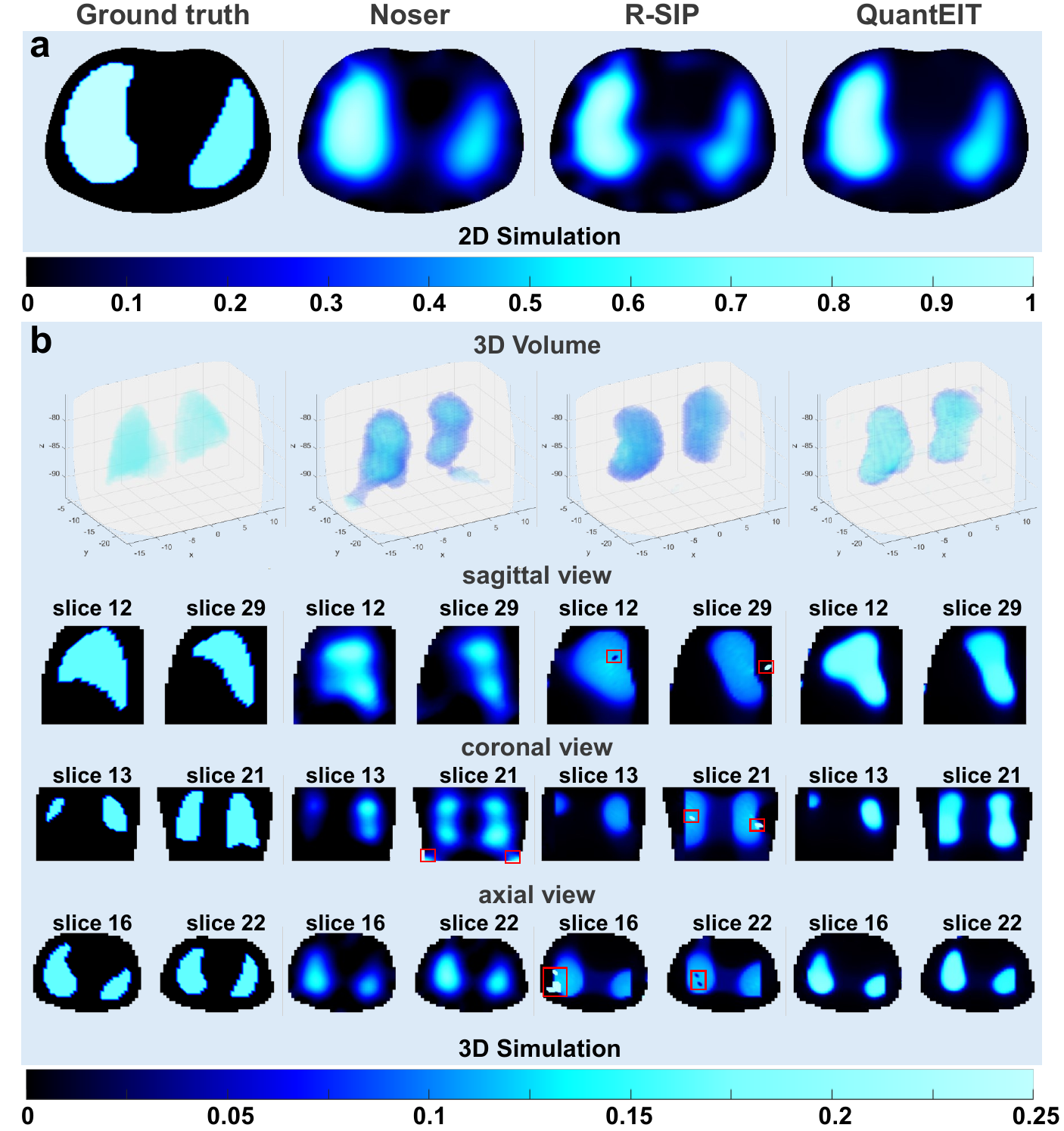}}
	\caption{Qualitative comparison of 2D (a) and 3D (b) EIT simulation reconstruction results across different algorithms. Red boxes highlight regions with apparent artifacts.}

	\label{qualitative_comparison}
\end{figure}

\begin{figure}
\centerline{\includegraphics[scale=0.37]{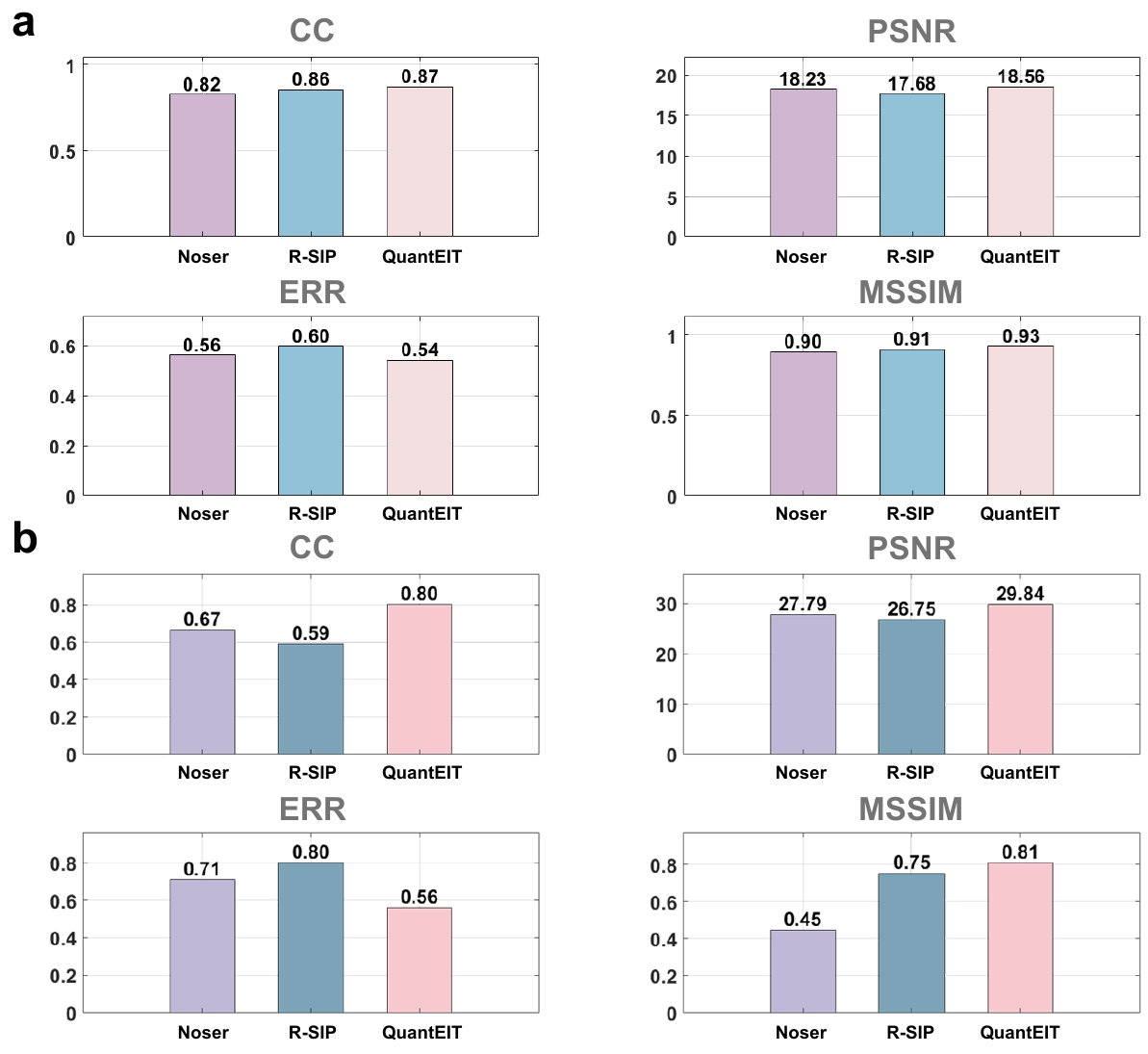}}
	\caption{Quantitative comparison of 2D (a) and 3D (b) EIT simulation reconstruction results across different algorithms.}

	\label{quantitative_comparison}
\end{figure}

To comprehensively evaluate the reconstruction performance, we adopted two protocols commonly used in EIT studies. For 2D simulations, max normalization was applied to the reconstructed conductivity distributions to emphasize the relative contrast and spatial variation, which is particularly relevant given the inherent scaling ambiguity in 2D inverse problems. In contrast, for 3D simulations, raw (unnormalized) reconstructions were assessed as a complementary analysis to evaluate the capability of each method to recover the magnitude of conductivity changes.

Fig.~\ref{qualitative_comparison} presents qualitative reconstruction results for both simulation scenarios. In the 2D case (Fig.~\ref{qualitative_comparison}a), all three methods correctly reconstructed distinguishable bilateral lung contours and accurately captured the conductivity differences reflecting the non-ventilation condition in the left lung. However, Noser exhibited limited fidelity in recovering the detailed lung structures, while both R-SIP and QuantEIT produced clearer boundary delineation. Notably, QuantEIT achieved superior structure preservation, along with a cleaner background and reduced artifacts. For the 3D case (Fig.~\ref{qualitative_comparison}b), volumetric reconstructions and representative slices along the x-, y-, and z-axes are displayed. All three methods successfully reconstructed distinguishable bilateral lung regions. Compared to Noser, R-SIP yielded better structural delineation but suffered from pronounced spurious noise and less accurate conductivity values. QuantEIT achieved the most accurate and visually coherent reconstruction, exhibiting the clearest lung contours, the most faithful conductivity levels, and substantially cleaner background with fewer artifacts.

Quantitative evaluations are summarized in Fig.~\ref{quantitative_comparison}. For 2D simulations (Fig.~\ref{quantitative_comparison}a), QuantEIT achieved the highest CC (0.87), PSNR (18.56 dB), and MSSIM (0.93), as well as the lowest ERR (0.54), outperforming Noser and R-SIP across all metrics. In 3D simulations (Fig.~\ref{quantitative_comparison}b), QuantEIT again exhibited superior quantitative performance with a CC of 0.80, PSNR of 29.84 dB, and MSSIM of 0.81, alongside a markedly lower error of 0.56 compared to other methods. These improvements indicate that QuantEIT not only enhances relative spatial contrast but also achieves more accurate absolute conductivity reconstruction in 3D volumes.

Overall, the assessments consistently demonstrate the effectiveness of QuantEIT in recovering both the structural and numerical characteristics of 2D and 3D chest conductivity distributions.

\subsubsection{Ablation Study}

\begin{figure}
\centerline{\includegraphics[scale=0.435]{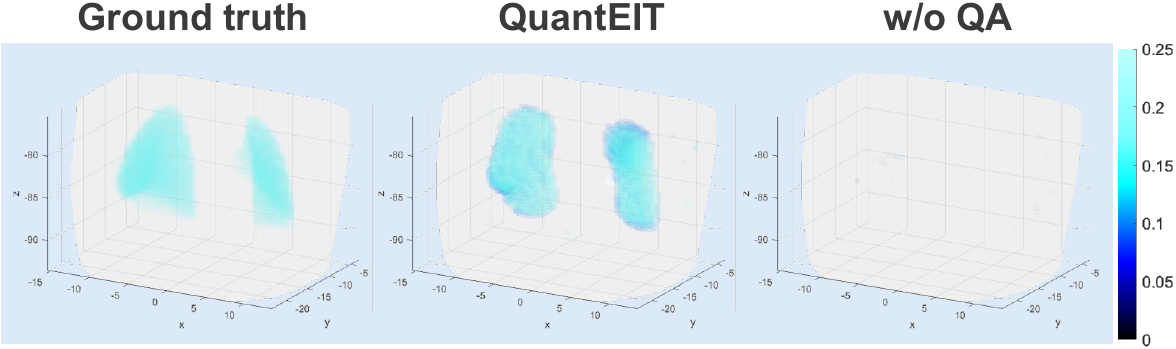}}
	\caption{Ablation study of the proposed QuantEIT on 3D simulation.}

	\label{ablation}
\end{figure}

To further assess the contribution of the quantum-assisted latent embedding, we conducted an ablation study on the 3D simulation. The outputs of the quantum circuits were removed and replaced with vectors of the same dimensionality. We considered two variants: (i) constant vectors filled with ones and (ii) learnable latent vectors initialized randomly and optimized during training. As shown in Fig.~\ref{ablation}, both configurations without quantum assistance failed to reconstruct any meaningful 3D structures, regardless of whether the latent representation was fixed or learned. In contrast, the proposed QuantEIT model, which employs quantum circuits to generate expressive non-linear latent inputs, successfully recovered the anatomical regions of the bilateral lungs with clear contours and consistent conductivity values closely matching the ground truth. These results demonstrate that the inherent entanglement and non-linearity of the quantum-assisted latent space provide essential prior information that cannot be substituted by either static or trainable embeddings.

\subsubsection{Computational Complexity and Inference Time}

\begin{table}[t]
\centering
\caption{Computational complexity and inference time comparison}
\label{tab:complexity}
\begin{tabular}{lcccc}
\toprule
\textbf{Method} & \textbf{Task} & \textbf{\# Parameters} & \textbf{FLOPs} \\
\midrule
QuantEIT & 2D & 20,484 & Negligible \\
QuantEIT & 3D & 204,804 & Negligible  \\
R-SIP & 2D & 8,854,096 & 17.72M  \\
R-SIP & 3D & 82,618,960 & 165.24M  \\
\bottomrule
\end{tabular}
\end{table}

\begin{figure}
\centerline{\includegraphics[scale=0.3]{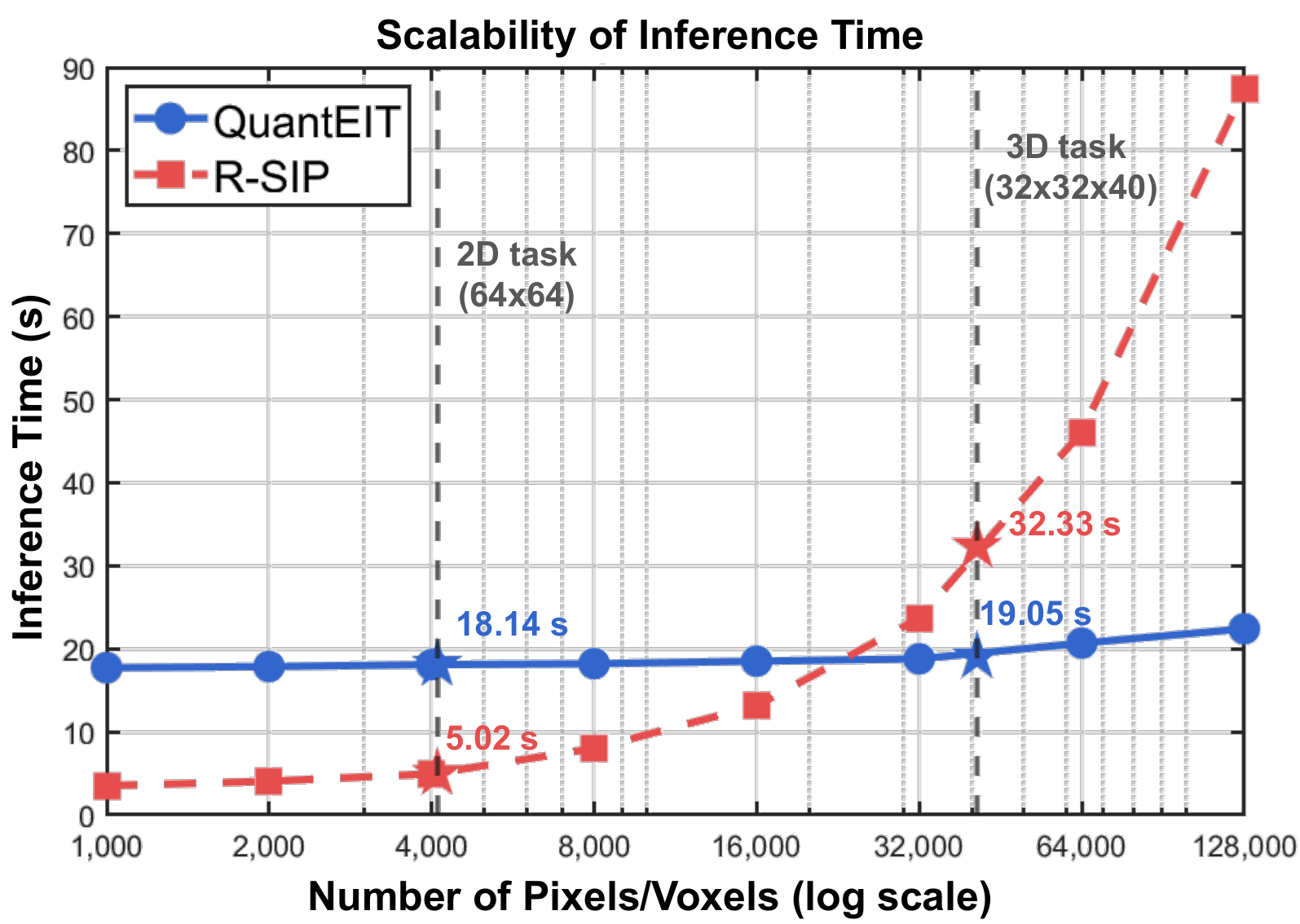}}
	\caption{Scalability of inference time with respect to the number of pixels/voxels (log scale).}

	\label{inference_time}
\end{figure}

We quantitatively compared the computational complexity and inference time of the proposed QuantEIT and the baseline R-SIP across 2D and 3D reconstruction tasks. Table~\ref{tab:complexity} summarizes the computational complexity of QuantEIT and R-SIP under different reconstruction scenarios. As summarized in Table~\ref{tab:complexity}, QuantEIT exhibits a substantially smaller number of learnable parameters due to the use of a compact quantum-assisted latent representation and a single linear mapping layer. Specifically, QuantEIT contains only 0.02M parameters for 2D reconstruction and 0.20M parameters for 3D reconstruction, while R-SIP requires 8.85M and 82.62M parameters, respectively.

In terms of floating-point operations (FLOPs), R-SIP involves 17.72 million FLOPs for 2D reconstruction and 165.24 million FLOPs for 3D reconstruction. In contrast, QuantEIT does not rely on deep feedforward layers and therefore has negligible FLOPs in the classical computation path, aside from the single linear projection. 

Despite its lightweight architecture, the inference time of QuantEIT is primarily bottlenecked by the CPU-based simulation of quantum circuits. As shown in Fig.~\ref{inference_time}, QuantEIT achieves nearly constant runtime across increasing output dimensions—18.14 seconds for 2D tasks ($64\times64$) and 19.05 seconds for 3D tasks ($32\times32\times40$). This stability stems from its fixed-size quantum circuit, whose simulation cost dominates but does not scale with the number of pixels or voxels. In contrast, R-SIP is faster in 2D (5.02 seconds) but suffers from rapidly increasing inference time at higher resolutions, reaching 32.33 seconds in 3D and nearly 90 seconds at the highest tested resolution. This degradation is attributed to its fully connected layers, whose computational complexity scales poorly in high-dimensional settings.

\begin{figure}
\centerline{\includegraphics[scale=0.46]{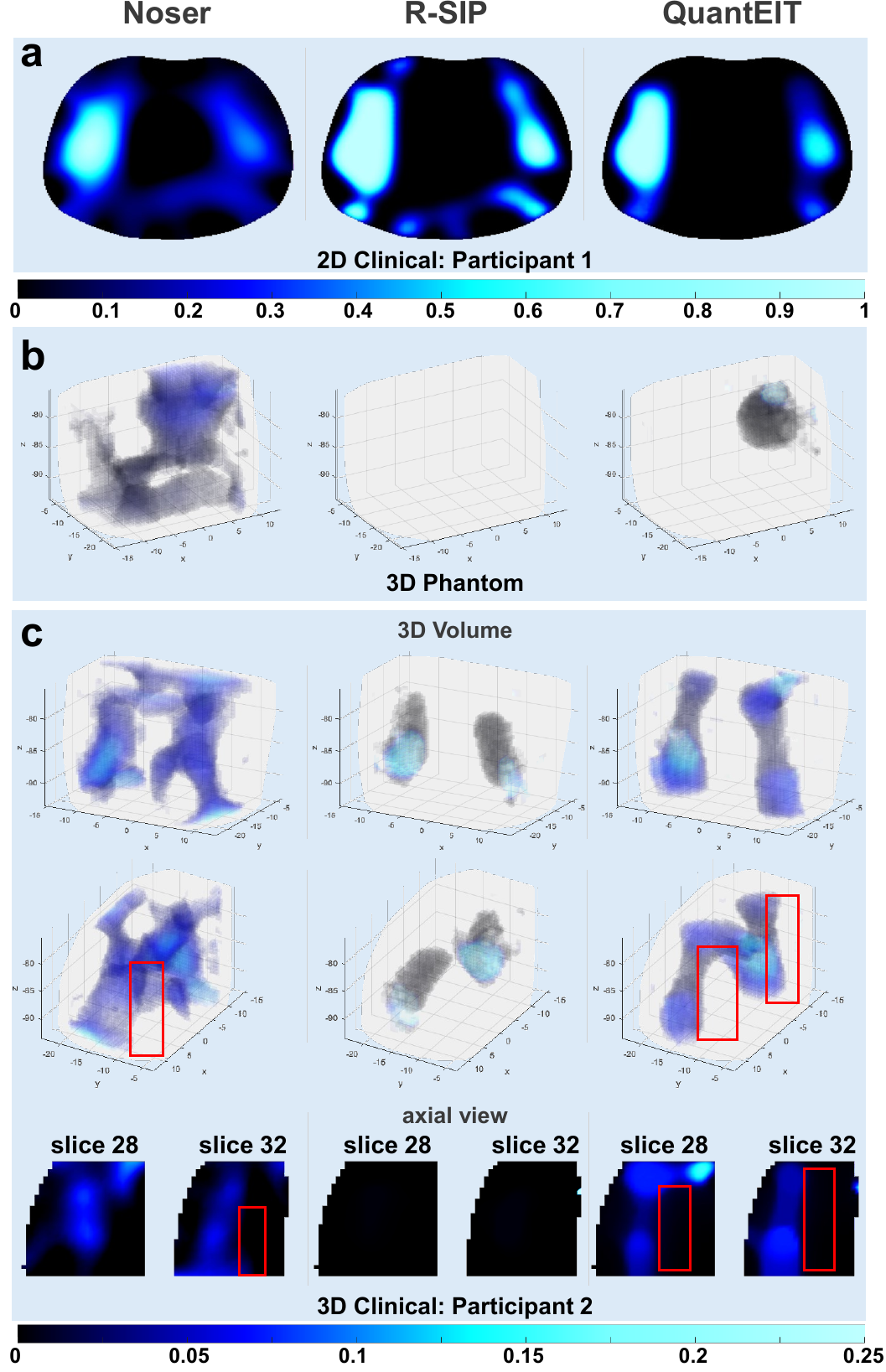}}
	\caption{Qualitative comparison of EIT reconstruction results across different algorithms on (a) 2D clinical data  (Participant 1), (b) 3D phantom data, and (c) 3D clinical data (Participant 2).}

	\label{qualitative_comparison_realworld}
\end{figure}

\begin{figure*}
\centerline{\includegraphics[scale=0.42]{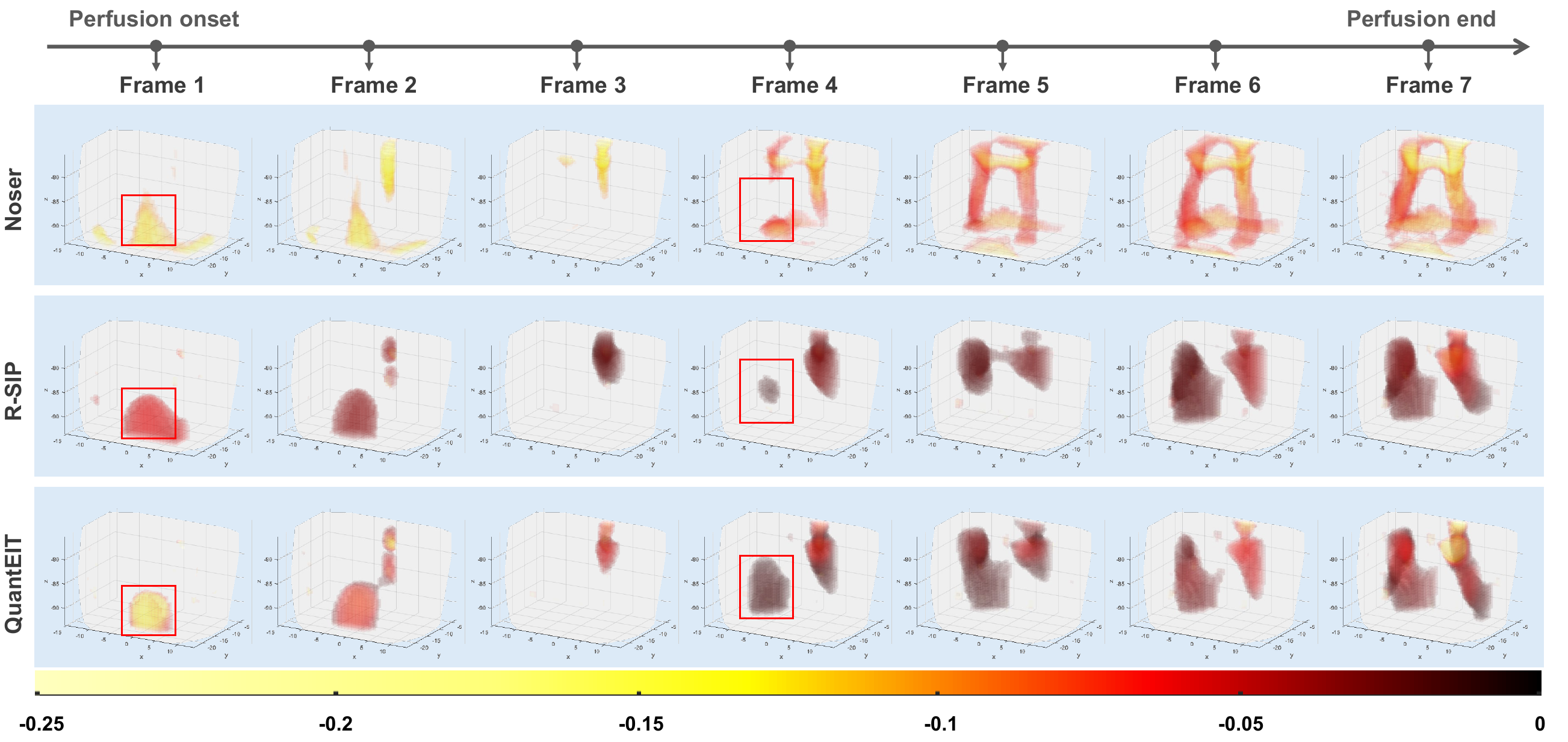}}
	\caption{Qualitative comparison of EIT reconstruction results across different algorithms on 3D clinical data from participant 3.}

	\label{qualitative_comparison_realworld2}
\end{figure*}

These results highlight QuantEIT’s strong scalability and resolution-agnostic inference characteristics, making it particularly suitable for high-dimensional EIT tasks. All experiments were conducted on a laptop with an Intel Core i9-14900HX CPU and NVIDIA RTX 4090 GPU, using PennyLane to simulate quantum circuits in CPU mode.

Notably, further acceleration is possible if quantum simulation is offloaded to GPU or executed on quantum hardware. As PennyLane currently performs simulation on the CPU by default, particularly on standard Windows machines, future support for GPU backends such as those available on Linux systems with NVIDIA cuQuantum, or support for native quantum execution, could significantly reduce runtime.

\subsection{Phantom Experiments}
Fig.~\ref{qualitative_comparison_realworld}b presents the 3D reconstruction results on the phantom dataset. The traditional Noser algorithm produces noisy and spatially diffuse structures, making it difficult to localize the ventilated region. R-SIP, limited by its shallow network architecture, fails to recover meaningful conductivity changes, resulting in a nearly blank reconstruction. In contrast, the proposed QuantEIT successfully reconstructs a compact and localized region of conductivity increase, which aligns with the left lung sac being ventilated. This demonstrates its superior capability in detecting subtle conductivity variations under weak signal conditions in real-world scenarios.

\subsection{Clinical Experiments}



To further evaluate the clinical applicability of the proposed approach, we conducted experiments on real patient data. Fig.~\ref{qualitative_comparison_realworld}a illustrates the 2D reconstruction results of Participant 1. The proposed QuantEIT clearly reveals the non-ventilation region in the left lung, which is consistent with the clinical reference provided by the widely used Noser algorithm. Both R-SIP and QuantEIT produce sharper contour boundaries compared to Noser; however, QuantEIT demonstrates superior noise robustness and a cleaner background.

Fig.\ref{qualitative_comparison_realworld}c shows the 3D reconstruction results for Participant 2. To provide anatomical reference, a CT scan was performed, as shown in Fig.\ref{CT_clinical}. Specifically, Fig.\ref{CT_clinical}a and Fig.\ref{CT_clinical}b reveal partial posterior lung volume loss (highlighted by red boxes), while the anterior lung regions remain structurally intact, indicating localized dorsal lung collapse. This clinically confirmed abnormality serves as a radiological benchmark for evaluating reconstruction performance. Among the three methods, R-SIP struggles to capture subtle conductivity variations due to its shallow MLP architecture, resulting in sparse and anatomically implausible reconstructions. Noser roughly indicates the left posterior defect but suffers from noise artifacts and lacks spatial continuity. In contrast, the proposed QuantEIT achieves the most faithful reconstruction: as highlighted by the red boxes in Fig.~\ref{qualitative_comparison_realworld}c, it accurately delineates the bilateral lung contours and clearly recovers the posterior ventilation defects, closely matching the CT findings. These results demonstrate that QuantEIT provides both robust noise suppression and anatomically consistent reconstruction under clinically realistic conditions.

Fig.~\ref{qualitative_comparison_realworld2} presents a qualitative comparison of 3D EIT reconstruction results from Participant 3 during the pulmonary perfusion phase (Frame 1 to Frame 7), across three algorithms. At Frame 1 (perfusion onset), impedance changes first appear near the cardiac region. The injected saline then gradually diffuses into both lungs by Frame 7. Noser captures early changes but produces blurry results. R-SIP shows weaker conductivity variation at Frame 1 and almost fails to reconstruct the perfusion signal at Frame 3. QuantEIT yields the most consistent and anatomically coherent results across all frames, accurately tracking the spatiotemporal perfusion process.

\section{Conclusion}
We proposed QuantEIT, an ultra-lightweight quantum-assisted inference framework for EIT reconstruction. By combining parallel 2-qubit quantum circuits with a single linear mapping layer, QuantEIT significantly reduces model complexity while maintaining high reconstruction quality. Extensive experiments on simulated and clinical 2D and 3D EIT lung imaging data demonstrated that QuantEIT consistently outperforms both traditional methods, such as Noser, and SOTA unsupervised approaches like R-SIP, achieving superior structural fidelity and noise robustness with only 0.2\% of the parameters. These results underline the potential of hybrid quantum–classical models as practical and efficient solutions for ill-posed inverse problems. Importantly, the proposed approach can be readily deployed on conventional hardware via quantum simulation and provides a clear pathway for future extension to quantum hardware implementations. Future research will focus on scaling the approach to other reconstruction tasks such as CT and Positron Emission Tomography, exploring richer quantum circuit designs, and validating clinical applicability across broader scenarios.

\FloatBarrier
\bibliographystyle{IEEEtran}
\bibliography{reference}

\begin{thebibliography}{10}
\providecommand{\url}[1]{#1}
\csname url@samestyle\endcsname
\providecommand{\newblock}{\relax}
\providecommand{\bibinfo}[2]{#2}
\providecommand{\BIBentrySTDinterwordspacing}{\spaceskip=0pt\relax}
\providecommand{\BIBentryALTinterwordstretchfactor}{4}
\providecommand{\BIBentryALTinterwordspacing}{\spaceskip=\fontdimen2\font plus
\BIBentryALTinterwordstretchfactor\fontdimen3\font minus \fontdimen4\font\relax}
\providecommand{\BIBforeignlanguage}[2]{{%
\expandafter\ifx\csname l@#1\endcsname\relax
\typeout{** WARNING: IEEEtran.bst: No hyphenation pattern has been}%
\typeout{** loaded for the language `#1'. Using the pattern for}%
\typeout{** the default language instead.}%
\else
\language=\csname l@#1\endcsname
\fi
#2}}
\providecommand{\BIBdecl}{\relax}
\BIBdecl

\bibitem{b2}
P.~Metherall, D.~C. Barber, R.~H. Smallwood, and B.~H. Brown, ``Three-dimensional electrical impedance tomography,'' \emph{Nature}, vol. 380, no. 6574, pp. 509--512, 1996.

\bibitem{b1}
H.~Fang, R.~B. Liu, J.~Sun, P.~O. Bagnaninchi, Z.~Liu, and Y.~Yang, ``Multi-modal eit imaging using lensfree and flexible impedance sensor,'' in \emph{2024 IEEE SENSORS}.\hskip 1em plus 0.5em minus 0.4em\relax IEEE, 2024, pp. 1--4.

\bibitem{b3}
I.~Frerichs, J.~Hinz, P.~Herrmann, G.~Weisser, G.~Hahn, M.~Quintel, and G.~Hellige, ``Regional lung perfusion as determined by electrical impedance tomography in comparison with electron beam ct imaging,'' \emph{IEEE transactions on medical imaging}, vol.~21, no.~6, pp. 646--652, 2002.

\bibitem{b4}
L.~Eichler, J.~Mueller, J.~Grensemann, I.~Frerichs, C.~Z{\"o}llner, and S.~Kluge, ``Lung aeration and ventilation after percutaneous tracheotomy measured by electrical impedance tomography in non-hypoxemic critically ill patients: a prospective observational study,'' \emph{Annals of intensive care}, vol.~8, pp. 1--8, 2018.

\bibitem{b5}
K.~Zhang, R.~Guo, M.~Li, F.~Yang, S.~Xu, and A.~Abubakar, ``Supervised descent learning for thoracic electrical impedance tomography,'' \emph{IEEE Transactions on Biomedical Engineering}, vol.~68, no.~4, pp. 1360--1369, 2020.

\bibitem{b6}
X.-Y. Ke, W.~Hou, Q.~Huang, X.~Hou, X.-Y. Bao, W.-X. Kong, C.-X. Li, Y.-Q. Qiu, S.-Y. Hu, and L.-H. Dong, ``Advances in electrical impedance tomography-based brain imaging,'' \emph{Military Medical Research}, vol.~9, no.~1, p.~10, 2022.

\bibitem{b7}
Y.~Jiang and M.~Soleimani, ``Capacitively coupled electrical impedance tomography for brain imaging,'' \emph{IEEE transactions on medical imaging}, vol.~38, no.~9, pp. 2104--2113, 2019.

\bibitem{b8}
K.~Y. Aristovich, B.~C. Packham, H.~Koo, G.~S. Dos~Santos, A.~McEvoy, and D.~S. Holder, ``Imaging fast electrical activity in the brain with electrical impedance tomography,'' \emph{NeuroImage}, vol. 124, pp. 204--213, 2016.

\bibitem{b9}
S.~Wang, D.~Hu, M.~Zhang, J.~Qiu~Lin, W.~Chen, F.~Giorgio-Serchi, L.~Peng, Y.~Li, and Y.~Yang, ``A digital twin of electrical tomography for quantitative multiphase flow imaging,'' \emph{Communications Engineering}, vol.~1, no.~1, p.~41, 2022.

\bibitem{b10}
S.-N. Wang, F.~Giorgio-Serchi, and Y.-J. Yang, ``A virtual platform of electrical tomography for multiphase flow imaging,'' \emph{Physics of Fluids}, vol.~34, no.~10, 2022.

\bibitem{b11}
M.~Vauhkonen, D.~Vad{\'a}sz, P.~A. Karjalainen, E.~Somersalo, and J.~P. Kaipio, ``Tikhonov regularization and prior information in electrical impedance tomography,'' \emph{IEEE transactions on medical imaging}, vol.~17, no.~2, pp. 285--293, 1998.

\bibitem{b22}
M.~Cheney, D.~Isaacson, J.~C. Newell, S.~Simske, and J.~Goble, ``Noser: An algorithm for solving the inverse conductivity problem,'' \emph{International Journal of Imaging systems and technology}, vol.~2, no.~2, pp. 66--75, 1990.

\bibitem{b12}
J.~Li, S.~Yue, M.~Ding, Z.~Cui, and H.~Wang, ``Adaptive $ l\_ $\{$p$\}$ $ regularization for electrical impedance tomography,'' \emph{IEEE Sensors Journal}, vol.~19, no.~24, pp. 12\,297--12\,305, 2019.

\bibitem{b13}
A.~Borsic, B.~M. Graham, A.~Adler, and W.~R. Lionheart, ``In vivo impedance imaging with total variation regularization,'' \emph{IEEE transactions on medical imaging}, vol.~29, no.~1, pp. 44--54, 2009.

\bibitem{b14}
H.~K. Aggarwal, M.~P. Mani, and M.~Jacob, ``Modl: Model-based deep learning architecture for inverse problems,'' \emph{IEEE transactions on medical imaging}, vol.~38, no.~2, pp. 394--405, 2018.

\bibitem{b16}
J.~Xiang, Y.~Dong, and Y.~Yang, ``Fista-net: Learning a fast iterative shrinkage thresholding network for inverse problems in imaging,'' \emph{IEEE Transactions on Medical Imaging}, vol.~40, no.~5, pp. 1329--1339, 2021.

\bibitem{b38}
Z.~Chen, H.~Zhang, D.~Hu, C.~Tan, Z.~Liu, and Y.~Yang, ``Point-cloud transformer for 3-d electrical impedance tomography,'' \emph{IEEE Transactions on Instrumentation Measurement}, vol.~73, p. 3413161, 2024.

\bibitem{b46}
Z.~Liu, Z.~Chen, and Y.~Yang, ``Review of machine learning for bioimpedance tomography in regenerative medicine,'' in \emph{Diverse Perspectives and State-of-the-Art Approaches to the Utilization of Data-Driven Clinical Decision Support Systems}.\hskip 1em plus 0.5em minus 0.4em\relax IGI Global Scientific Publishing, 2023, pp. 271--292.

\bibitem{b23}
S.~J. Hamilton and A.~Hauptmann, ``Deep d-bar: Real-time electrical impedance tomography imaging with deep neural networks,'' \emph{IEEE transactions on medical imaging}, vol.~37, no.~10, pp. 2367--2377, 2018.

\bibitem{b24}
C.~Tan, S.~Lv, F.~Dong, and M.~Takei, ``Image reconstruction based on convolutional neural network for electrical resistance tomography,'' \emph{IEEE Sensors Journal}, vol.~19, no.~1, pp. 196--204, 2018.

\bibitem{b25}
F.~Li, C.~Tan, and F.~Dong, ``Electrical resistance tomography image reconstruction with densely connected convolutional neural network,'' \emph{IEEE Transactions on Instrumentation and Measurement}, vol.~70, pp. 1--11, 2020.

\bibitem{b26}
J.~K. Seo, K.~C. Kim, A.~Jargal, K.~Lee, and B.~Harrach, ``A learning-based method for solving ill-posed nonlinear inverse problems: A simulation study of lung eit,'' \emph{SIAM journal on Imaging Sciences}, vol.~12, no.~3, pp. 1275--1295, 2019.

\bibitem{b27}
Y.~Chen, K.~Li, and Y.~Han, ``Electrical resistance tomography with conditional generative adversarial networks,'' \emph{Measurement Science and Technology}, vol.~31, no.~5, p. 055401, 2020.

\bibitem{b30}
H.~Fang, Z.~Liu, Y.~Feng, Z.~Qiu, P.~Bagnaninchi, and Y.~Yang, ``Multi-frequency electrical impedance tomography reconstruction with multi-branch attention image prior,'' \emph{arXiv preprint arXiv:2409.10794}, 2024.

\bibitem{b31}
Z.~Chen, J.~Xiang, P.-O. Bagnaninchi, and Y.~Yang, ``Mmv-net: A multiple measurement vector network for multifrequency electrical impedance tomography,'' \emph{IEEE Transactions on Neural Networks and Learning Systems}, vol.~34, no.~11, pp. 8938--8949, 2022.

\bibitem{b21}
D.~Ulyanov, A.~Vedaldi, and V.~Lempitsky, ``Deep image prior,'' in \emph{Proceedings of the IEEE conference on computer vision and pattern recognition}, 2018, pp. 9446--9454.

\bibitem{b28}
D.~Liu, J.~Wang, Q.~Shan, D.~Smyl, J.~Deng, and J.~Du, ``Deepeit: Deep image prior enabled electrical impedance tomography,'' \emph{IEEE Transactions on Pattern Analysis and Machine Intelligence}, vol.~45, no.~8, pp. 9627--9638, 2023.

\bibitem{b29}
Z.~Liu, Z.~Chen, H.~Fang, Q.~Wang, S.~Zhang, and Y.~Yang, ``Regularized shallow image prior for electrical impedance tomography,'' \emph{IEEE Transactions on Instrumentation and Measurement}, 2025.

\bibitem{b32}
S.~Chen, J.~Cotler, H.-Y. Huang, and J.~Li, ``The complexity of nisq,'' \emph{Nature Communications}, vol.~14, no.~1, p. 6001, 2023.

\bibitem{b33}
X.~Pan, Z.~Lu, W.~Wang, Z.~Hua, Y.~Xu, W.~Li, W.~Cai, X.~Li, H.~Wang, Y.-P. Song \emph{et~al.}, ``Deep quantum neural networks on a superconducting processor,'' \emph{Nature Communications}, vol.~14, no.~1, p. 4006, 2023.

\bibitem{b34}
T.~Ahmed, M.~Kashif, A.~Marchisio, and M.~Shafique, ``Quantum neural networks: A comparative analysis and noise robustness evaluation,'' \emph{arXiv preprint arXiv:2501.14412}, 2025.

\bibitem{b35}
T.~Hur, L.~Kim, and D.~K. Park, ``Quantum convolutional neural network for classical data classification,'' \emph{Quantum Machine Intelligence}, vol.~4, no.~1, p.~3, 2022.

\bibitem{b36}
M.~C. Caro, H.-Y. Huang, M.~Cerezo, K.~Sharma, A.~Sornborger, L.~Cincio, and P.~J. Coles, ``Generalization in quantum machine learning from few training data,'' \emph{Nature communications}, vol.~13, no.~1, p. 4919, 2022.

\bibitem{b37}
V.~Bergholm, J.~Izaac, M.~Schuld, C.~Gogolin, S.~Ahmed, V.~Ajith, M.~S. Alam, G.~Alonso-Linaje, B.~AkashNarayanan, A.~Asadi \emph{et~al.}, ``Pennylane: Automatic differentiation of hybrid quantum-classical computations,'' \emph{arXiv preprint arXiv:1811.04968}, 2018.

\bibitem{b41}
M.~A. Nielsen and I.~L. Chuang, \emph{Quantum computation and quantum information}.\hskip 1em plus 0.5em minus 0.4em\relax Cambridge university press, 2010.

\bibitem{b40}
D.~P. Kingma, ``Adam: A method for stochastic optimization,'' \emph{arXiv preprint arXiv:1412.6980}, 2014.

\bibitem{b43}
P.~C. Hansen, \emph{Rank-deficient and discrete ill-posed problems: numerical aspects of linear inversion}.\hskip 1em plus 0.5em minus 0.4em\relax SIAM, 1998.

\bibitem{b42}
L.~I. Rudin, S.~Osher, and E.~Fatemi, ``Nonlinear total variation based noise removal algorithms,'' \emph{Physica D: nonlinear phenomena}, vol.~60, no. 1-4, pp. 259--268, 1992.

\bibitem{b44}
A.~Adler and W.~R. Lionheart, ``Eidors: Towards a community-based extensible software base for eit,'' in \emph{Proceedings of the 6th Conference on Biomedical Applications of Electrical Impedance Tomography, London}, 2005.

\bibitem{b45}
Z.~Wang, A.~C. Bovik, H.~R. Sheikh, and E.~P. Simoncelli, ``Image quality assessment: from error visibility to structural similarity,'' \emph{IEEE transactions on image processing}, vol.~13, no.~4, pp. 600--612, 2004.

\end{thebibliography}

\end{document}